\documentclass[sigconf]{acmart}
\AtBeginDocument{%
  \providecommand\BibTeX{{%
    \normalfont B\kern-0.5em{\scshape i\kern-0.25em b}\kern-0.8em\TeX}}}

\setcopyright{acmcopyright}
\copyrightyear{2018}
\acmYear{2018}
\acmDOI{10.1145/1122445.1122456}

\acmConference[Woodstock '18]{Woodstock '18: ACM Symposium on Neural
  Gaze Detection}{June 03--05, 2018}{Woodstock, NY}
\acmBooktitle{Woodstock '18: ACM Symposium on Neural Gaze Detection,
  June 03--05, 2018, Woodstock, NY}
\acmPrice{15.00}
\acmISBN{978-1-4503-XXXX-X/18/06}

\usepackage{graphicx}
\usepackage{subfigure}
\usepackage{algorithmic}
\usepackage{hyperref}
\usepackage{xspace}
\usepackage{float}
\newcommand{\model}{GIPA\xspace}

\begin{document}

\title{GIPA: General Information Propagation Algorithm for Graph Learning}

%

\author{Qinkai Zheng$^{\dagger*}$, Houyi Li$^{\dagger*}$, Peng Zhang$^\ddagger$,\\Zhixiong Yang$^\dagger$, Guowei Zhang$^\dagger$, Xintan Zeng$^\dagger$, Yongchao Liu$^\dagger$}
\affiliation{
 \institution{$^\dagger$Ant Group}
 \institution{$^\ddagger$Alibaba Group}
 \city{Hangzhou}
 \country{China}
}
\email{{qinkai.zheng1028, zhangpeng04}@gmail.com, {houyi.lhy, zhixiong.yangzx, bale.zgw, xintan.zxt, yongchao.ly}@antgroup.com}
\thanks{$^*$Equal contributions to this research.}








\renewcommand{\shortauthors}{Zheng and Li, et al.}

\begin{abstract}
Graph neural networks (GNNs) have been popularly used in analyzing graph-structured data,  showing promising results in various applications such as node classification, link prediction and network recommendation. 
In this paper, we present a new graph attention neural network, namely \model, for attributed graph data learning.
\model consists of three key components: attention, feature propagation and aggregation.
Specifically, the attention component introduces a new  multi-layer perceptron based multi-head to generate better  non-linear feature mapping and representation than conventional implementations such as dot-product. 
The propagation component considers not only node features but also edge features, which differs from existing GNNs that merely consider node features.
The aggregation component uses a residual connection to generate the final embedding.
We evaluate the performance of \model using the Open Graph Benchmark proteins (ogbn-proteins for short) dataset.
The experimental results reveal that \model can beat the  state-of-the-art models in terms of prediction accuracy, e.g., \model achieves an average test ROC-AUC of $0.8700\pm 0.0010$ and outperforms all the previous methods listed in the ogbn-proteins leaderboard.
\end{abstract}



\maketitle
\pagestyle{plain}

\section{Introduction}
\label{sec:intro}
Graph representation learning typically aims to learn an   informative embedding for each graph node based on the graph topology (link) information.
Generally, the embedding of a node is represented as a low-dimensional feature vector, which can be used to facilitate downstream applications.
This research starts from homogeneous graphs that have only one type of nodes and one type of edges. The purpose is to learn node representations from the graph topology~\cite{grover2016node2vec, perozzi2014deepwalk, dai2016discriminative}.
Specifically, given a node $u$,  either breadth-first search, depth-first search or random walks is used to identify a set of neighboring nodes. Then, the $u$'s embedding is learnt by maximizing the co-occurrence probability of $u$ and its neighbors.
In reality, nodes and edges can carry a rich set of information such as attributes, texts, images or videos, beyond graph structure.
These information can be used to generate node or edge feature by means of a feature transformation function, which  further improves the performance of network embedding~\cite{liao2018attributed}.

These pioneer studies on graph embedding have limited capability to capturing neighboring information from a graph because they are based on shallow learning models such as SkipGram~\cite{mikolov2013distributed}.
Moreover, transductive learning is used in these graph embedding methods which cannot generalize to new nodes that are absent in the training graph.

On the other hand, graph neural networks~\cite{kipf2016semi, hamilton2017inductive, velivckovic2017graph} are proposed to overcome the limitations of graph embedding models.
GNNs employ deep neural networks to aggregate feature information from neighboring nodes and thereby have the potential to gain better aggregated embedding.
GNNs can support inductive learning and infer the class labels of unseen nodes during prediction~\cite{hamilton2017inductive, velivckovic2017graph}.
The success of GNNs is mainly based on the neighborhood information aggregation.
Two critical challenges to GNNs are \textit{which neighboring nodes of a target node are involved in message passing, and how much contribution each neighboring node makes to the aggregated embedding}.
For the former question, neighborhood sampling~\cite{hamilton2017inductive, ying2018graph, chen2018fastgcn, huang2018adaptive, zou2019layer, ji2020accelerating} is proposed for large dense or power-law graphs.
For the latter, neighbor importance estimation is used to add different weights to different neighboring nodes during feature propagation.
Importance sampling~\cite{chen2018fastgcn, zou2019layer, ji2020accelerating} and attention~\cite{velivckovic2017graph,liu2019geniepath, wang2019heterogeneous, yun2019graph, hu2020heterogeneous} are two popular techniques.

Importance sampling is a specialization of neighborhood sampling, where the importance weight of a neighboring node is drawn from a distribution over nodes.
This distribution can be derived from normalized Laplacian matrices~\cite{chen2018fastgcn, zou2019layer} or jointly learned with GNNs~\cite{ji2020accelerating}.
With this distribution, each step samples a subset of neighbors and aggregates their information with importance weights.
Similar to importance sampling, attention also adds importance weights to neighbors.
Nevertheless, attention differs from importance sampling.
Attention is represented as a neural network and always learned as a part of a GNN model.
In contrast, importance sampling algorithms use statistical models without trainable parameters.
Existing attention ranks nodes but does not drop off any of them. On the contrary, importance sampling reduces the number of neighbors.
Basically, we would expect an (almost equal to) zero importance score for noise neighbors and higher scores for strongly correlated neighbors.

\begin{figure*}[!h]
    \centering
    \includegraphics[width=\textwidth]{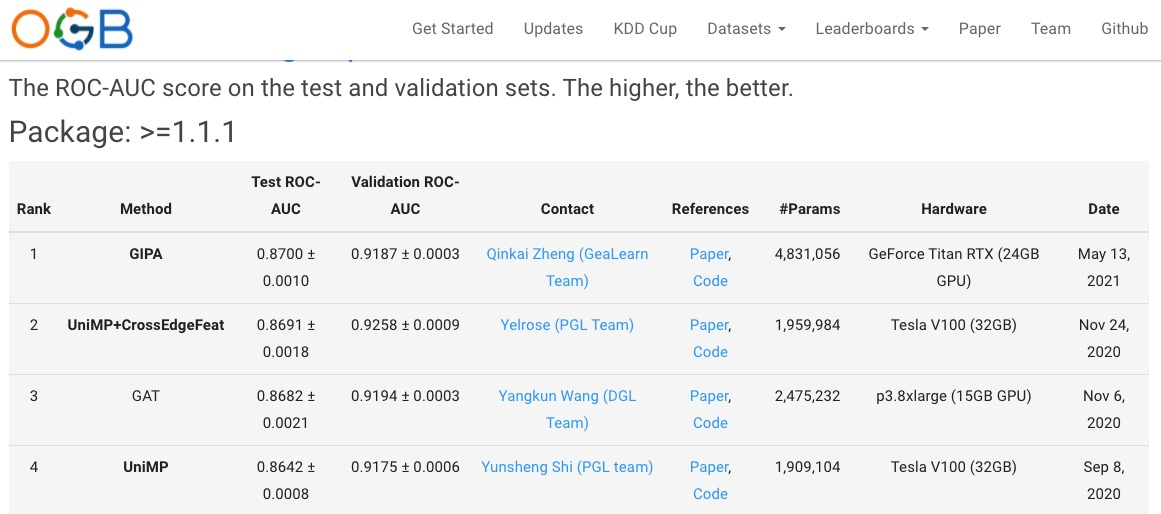}
    \caption{A snapshot of the leaderboard of ogbn-proteins with \model as the champion.}
    \label{fig:ogbn_proteins}
\end{figure*}

In this paper, we present a new graph attention neural network, namely GIPA (\underline{G}eneral \underline{I}nformation \underline{P}ropagation \underline{A}lgorithm), for attributed graphs.
In general, GIPA consists of three components, i.e.,  attention, propagation and aggregation.
The attention component uses a multi-head method which  is implemented with a multi-layer perceptron (MLP). 
The propagation component incorporates both node and edge features, unlike GAT~\cite{velivckovic2017graph} that uses only node features.
The aggregation component reduces messages from the neighbors of a given node $i$ and concatenates the resulted message with a linear projection of node $i$'s features by means of a residual connection.
Experiments on the Open Graph Benchmark (OGB)~\cite{hu2020open} proteins dataset (ogbn-proteins)  demonstrate that GIPA reaches the best accuracy with an average test ROC-AUC of $0.8700\pm 0.0010$ compared with the  state-of-the-art methods listed in the ogbs-proteins leaderboard~\footnote{https://ogb.stanford.edu/docs/leader\_nodeprop/\#ogbn-proteins}.
Up to date, this performance has been exhibited to the public in the leaderboard with \model in the first place, as shown in~\figurename~\ref{fig:ogbn_proteins}.

\section{Related Work}
\label{sec:related}
In this part, we survey existing attention primitive implementations in brief.
\cite{bahdanau2014neural} proposed an additive attention that calculates the attention alignment score using a simple feed-forward neural network with only one hidden layer.
The alignment score $score(q,k)$ between two vectors $q$ and $k$ is defined as
\begin{equation}
	score(q,k) = u^T\tanh(W[q;k]), 
\end{equation}
where $u$ is an attention vector and the attention weight $\alpha_{q,k}$ is computed by normalizing $score_{q,k}$ over all ${q,k}$ values with the softmax function.
The core of the additive attention lies in the use of attention vector $u$.
This idea has been widely adopted by several algorithms~\cite{yang2016hierarchical, pavlopoulos2017deeper} for neural language processing.
\cite{luong2015effective} introduces a global attention and a local attention.
In global attention, the alignment score can be computed by three alternatives: dot-product ($q^Tk$), general ($q^TWk$) and concat ($W[q;k]$).
In contrast, local attention computes the alignment score solely from a vector ($Wq$).
Likewise, both global and local attention normalize the alignment scores with the softmax function.
\cite{vaswani2017attention} proposed a self-attention mechanism based on scaled dot-products.
This self-attention computes the alignment scores between any $q$ and $k$ as follows.
\begin{equation}
	score_{q,k} = \frac{q^Tk}{\sqrt{d_k}}
\end{equation}
This attention differs from the dot-product attention~\cite{luong2015effective} by only a scaling factor of $\frac{1}{\sqrt{d_k}}$.
The scaling factor is used because the authors of~\cite{vaswani2017attention} suspected that for large values of $d_k$, the dot-products grow large in magnitude and thereby push the softmax function into regions where it has extremely small gradients.
Furthermore, a multi-head mechanism is proposed in order to stabilize the self-attention values computed.

In this paper, we introduce an MLP based multi-head implementation to compute attention scores.
In addition, these aforementioned attention primitives have been extended to use in heterogeneous graph models.
HAN~\cite{wang2019heterogeneous} uses a two-level hierarchical attention made out of a node-level attention and a semantic-level attention.
In HAN, the node-level attention learns the importance of a node to any other node in a meta-path, while the semantic-level one weighs all meta-paths.
HGT~\cite{zhang2019heterogeneous} weakens the dependency on meta-paths and instead uses meta-relation triplets as basic units.
In HGT, it uses node-type-aware feature transform functions and edge-type-aware multi-head attention to compute the importance of each edge to a target node.
It needs to be addressed that no evidence shows that heterogeneous models are always superior to homogeneous ones, and vice versa.

\section{Methodology}
\label{sec:method}
\subsection{Preliminaries}

\paragraph{Graph Neural Networks.} Consider an attributed graph $\mathcal{G}=\{\mathcal{V}, \mathcal{E}\}$, where $\mathcal{V}$ is the set of nodes and $\mathcal{E}$ is the set of edges. GNNs use the same model framework as follows~\cite{hamilton2017inductive,xu2018powerful}:

\begin{equation}
\label{eq:gnn}
h_i^k=\phi(h_i^{k-1}, \mathcal{F}_{agg}(\{h_j, j\in\mathcal{N}(i)\}))
\end{equation}
\noindent where $\mathcal{F}_{agg}$ represents an aggregation function and $\phi$ represents an update function. The objective of GNNs is to update the embedding of each node by aggregating the information from its neighbor nodes and the connections between them. 

\subsection{GIPA Architecture}

\begin{figure*}
    \centering
    \includegraphics[width=1.0\textwidth]{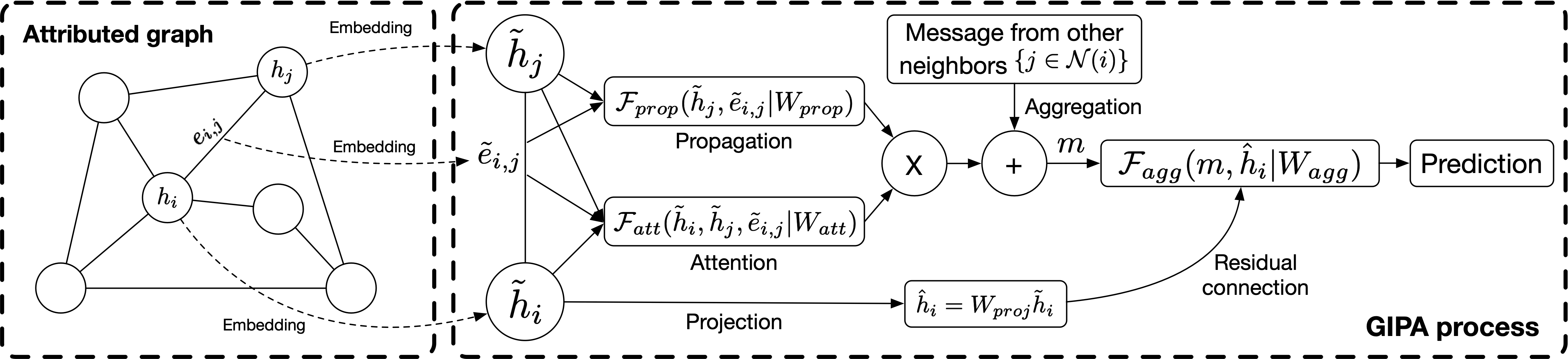}
    \caption{Process of \model, which consists of \textit{attention}, \textit{propagation}, and \textit{aggregation} process.}
    \label{fig:gipa_process}
\end{figure*}

In this part, we present the architecture of \model, which  extracts information from node features as well as edge features in a more general way. The process of \model is shown in~\figurename~\ref{fig:gipa_process}. Consider a node $i$ with feature $h_i$ and its neighbor nodes $j\in\mathcal{N}(i)$ with feature $h_j$. $e_{i, j}$ represents the edge feature between node $i$ and $j$. The problem is how to generate an expected embedding for node $i$ from its own node feature $h_i$, its neighbors' node features $\{h_j\}$ and the related edge features $\{e_{i, j}\}$. 

The \model process mainly consists of three parts: \texttt{attention}, \texttt{propagation} and \texttt{aggregation}. Firstly, \model computes the embedding $\tilde{h}_i$, $\tilde{h}_j$, and $\tilde{e}_{i, j}$ by means of linear projection from the features $h_i$, $h_j$ and $e_{i,j}$, respectively. Secondly, the \texttt{attention} process calculates the multi-head attention weights by using fully-connected layers on $\tilde{h}_i$, $\tilde{h}_j$, and $\tilde{e}_{i, j}$, respectively. Thirdly, the \texttt{propagation} process focuses on propagating information of each neighbor node $j$ by combining $j$'s node embedding $\tilde{h}_j$ with the associated edge embedding $\tilde{e}_{i, j}$. Finally, the \texttt{aggregation} process aggregates all messages from neighbors to update the embedding of $i$. The following subsections introduce the details of each process.

\subsubsection{Attention Process}

Different from the existing attention mechanisms like self-attention or scaled dot-product attention, we use MLP to realize a multi-head attention mechanism.
The \texttt{attention} process of \model can be formulated as follows:

\begin{equation}
\label{eq:att}
\tilde{a}_{i, j} = \mathcal{F}_{att}(\tilde{h}_i, \tilde{h}_j, \tilde{e}_{i, j}|W_{att}) = \text{MLP}([\tilde{h}_i || \tilde{h}_j || \tilde{e}_{i, j}]|W_{att})
\end{equation}

\noindent where the attention function $\mathcal{F}_{att}$ is realized by an \texttt{MLP} with learnable weights $W_{att}$ (without bias). Its input is the concatenation of the node embedding $\tilde{h}_i$ and $\tilde{h}_j$ as well as the edge embedding $\tilde{e}_{i, j}$. As the scale of parameters of \texttt{MLP} is adjustable, this attention mechanism could be more representative than previous ones simply based on dot-product. The final attention weight is calculated by an edge-wise \texttt{softmax} activation function:

\begin{equation}
\label{eq:softmax}
a_{i, j} = \text{softmax}(\tilde{a}_{i, j}) = \frac{\exp(\tilde{a}_{i, j})}{\sum_{k\in N(i)}\exp(\tilde{a}_{i, k})}
\end{equation}

\subsubsection{Propagation Process}

Unlike GAT~\cite{velivckovic2017graph} that only considers the node feature of neighbors, \model incorporates both node and edge features during the \textit{propagation} process:
\begin{equation}
\label{eq:prop}
p_{i, j} = \mathcal{F}_{prop}(\tilde{h}_j, \tilde{e}_{i,j}|W_{prop}) = \text{MLP}([\tilde{h}_j || \tilde{e}_{i, j}]|W_{prop}), 
\end{equation}
\noindent where the propagation function $\mathcal{F}_{prop}$ is also realized by an \text{MLP} with learnable weights $W_{prop}$. Its input is the concatenation of a neighbor node embedding $\tilde{h}_j$ and the related edge embedding $\tilde{e}_{i, j}$. Thus, the $propagation$ is done edge-wise rather than node-wise. 

Combining the results by \texttt{attention} and \texttt{propagation} by element-wise multiplication, \model gets the message $m_{i, j}$ of node $i$ from $j$:
\begin{equation}
\label{eq:msg}
m_{i, j} = a_{i, j} * p_{i, j}
\end{equation}

\subsubsection{Aggregation Process}

For each node $i$, \model repeats previous processes to get messages from its neighbors. The \texttt{aggregation} process first gathers all these messages by a reduce function, summation for example:

\begin{equation}
\label{eq:sum}
m_{i} = \sum_{j\in\mathcal{N}(i)}m_{i, j}
\end{equation}

\noindent Then, a residual connection between the linear projection $\hat{h}_i$ and the message of $m_i$ is added through concatenation: 

\begin{equation}
\label{eq:proj}
\hat{h}_i=W_{proj}\tilde{h}_i
\end{equation}

\begin{equation}
\label{eq:agg}
o_{i} = \mathcal{F}_{agg}(m, \hat{h}_i | W_{agg}) = \text{MLP}([\sum_{j\in\mathcal{N}(i)}m_{i, j} || \hat{h}_i]|W_{agg})
\end{equation}

\noindent where an \texttt{MLP} with learnable weights $W_{agg}$ is applied to get the final output $o_{i}$. 
Finally, we would like to emphasize that the process of \model can be easily extended to multi-layer variants by stacking the process multiple times.

\section{Experiments}
\label{sec:exp}
\subsection{Dataset and Settings}

\textbf{Dataset.} In our experiments, the \textit{ogbn-proteins} dataset from OGB~\cite{hu2020open} is used. The dataset is an undirected and weighted graph, containing 132,534 nodes of 8 different species and 79,122,504 edges with 8-dimensional features. The task is a multi-label binary classification problem with 112 classes representing different protein functions. 

\noindent \textbf{Evaluation metric.} The performance is measured by the average ROC-AUC scores. We follow the dataset splitting settings as recommended in OGB and test the performance of 10 different trained models with different random seeds. 

\noindent \textbf{Hyperparameters.} The hyperparamters used in \model and its training process are concluded in \tablename~\ref{tab:hyper}.

\noindent \textbf{Environment.} \model is implemented in Deep Graph Library (DGL)~\cite{wang2019deep} with Pytorch~\cite{paszke2019pytorch} as the backend. The experiments are done in a platform with GeForce TITAN RTX GPU (24GB RAM) and AMD Ryzen Threadripper 3960X CPU (24 Cores).

\begin{table}[t]
\caption{Hyperparameters of \model model.}
\begin{tabular}{lc}
\hline
\textbf{Hyperparameter}      & \textbf{Value} \\ \hline
Node embedding length        & 80             \\ 
Edge embedding length        & 16             \\ 
Number of attention layers   & 2              \\ 
Number of attention heads    & 8              \\ 
Number of propagation layers & 2              \\ 
Number of hidden units       & 80             \\ 
Number of GIPA layers        & 6              \\ 
Edge drop rate               & 0.1            \\ 
Aggregation                  & SUM            \\ 
Activation                   & ReLU           \\
Dropout of node embedding    & 0.1            \\
Dropout of attention         & 0.1            \\ 
Dropout of propagation       & 0.25           \\ 
Dropout of aggregation       & 0.25           \\ 
Dropout of the last FC layer & 0.5            \\ 
Learning rate & 0.01            \\ 
Optimizer & AdamW~\cite{loshchilov2017decoupled}  \\ \hline
\end{tabular}
\label{tab:hyper}
\end{table}
\begin{table}[!h]
\caption{Test and validation performance (ROC-AUC) on \textit{ogbn-proteins} dataset.}
\begin{tabular}{lcc}
\hline
Method              & Test ROC-AUC  & Validation ROC-AUC \\ \hline
MLP                 & $0.7204\pm0.0048$ & $0.7706\pm0.0014$      \\
GCN                 & $0.7251\pm0.0035$ & $0.7921\pm0.0018$      \\
GraphSAGE           & $0.7768\pm0.0020$ & $0.8334\pm0.0013$      \\
DeeperGCN           & $0.8496\pm0.0028$ & $0.8971\pm0.0011$      \\
GAT                 & $0.8682\pm0.0018$ & $0.9194\pm0.0003$      \\
UniMP               & $0.8642\pm0.0008$ & $0.9175\pm0.0006$      \\
UniMP+CrossEdgeFeat & $0.8691\pm0.0018$ & $\mathbf{0.9258\pm0.0009}$      \\
GIPA (ours)         & $\mathbf{0.8700\pm0.0010}$ & $0.9187\pm0.0003$      \\ \hline
\label{tab:performance}
\end{tabular}
\end{table}
\subsection{Performance}
\tablename~\ref{tab:performance} shows the average ROC-AUC and the standard deviation for the test set and the validation set, respectively. The results of the baselines are retrieved from the ogbn-proteins leaderboard\footnotemark[1]. Our \model outperforms all previous methods in the leaderboard and reaches an average test ROC-AUC higher than 0.8700 for the first time.

\section{Conclusion}
\label{sec:conclusion}
We have presented \model, a new graph attention network architecture for attributed graph data learning.
\model has investigated three technical novelties: an MLP based multi-head attention, a propagation function involving both node feature and edge feature, and a residual connection  aggregation method.
The performance evaluations on the ogbn-proteins dataset have demonstrated that \model yields an average test ROC-AUC of $0.8700\pm 0.0010$, showing the best results compared with the state-of-the-art methods listed in the ogbn-proteins leaderboard. In the future, we will extend \model to heterogeneous GNNs and conduct testing on the open graph datasets. 

\begin{acks}
This work is primarily presented by the GeaLearn team of Ant Group, China.
Qinkai Zheng was a research intern in the GeaLearn team.
We thank Dr. Changhua He and Dr. Wenguang Chen for their support and guidance.
\end{acks}

\bibliographystyle{ACM-Reference-Format}
\bibliography{reference}


\begin{thebibliography}{28}


\ifx \showCODEN    \undefined \def \showCODEN     #1{\unskip}     \fi
\ifx \showDOI      \undefined \def \showDOI       #1{#1}\fi
\ifx \showISBNx    \undefined \def \showISBNx     #1{\unskip}     \fi
\ifx \showISBNxiii \undefined \def \showISBNxiii  #1{\unskip}     \fi
\ifx \showISSN     \undefined \def \showISSN      #1{\unskip}     \fi
\ifx \showLCCN     \undefined \def \showLCCN      #1{\unskip}     \fi
\ifx \shownote     \undefined \def \shownote      #1{#1}          \fi
\ifx \showarticletitle \undefined \def \showarticletitle #1{#1}   \fi
\ifx \showURL      \undefined \def \showURL       {\relax}        \fi
\providecommand\bibfield[2]{#2}
\providecommand\bibinfo[2]{#2}
\providecommand\natexlab[1]{#1}
\providecommand\showeprint[2][]{arXiv:#2}

\bibitem[\protect\citeauthoryear{Bahdanau, Cho, and Bengio}{Bahdanau
  et~al\mbox{.}}{2015}]%
        {bahdanau2014neural}
\bibfield{author}{\bibinfo{person}{Dzmitry Bahdanau},
  \bibinfo{person}{Kyunghyun Cho}, {and} \bibinfo{person}{Yoshua Bengio}.}
  \bibinfo{year}{2015}\natexlab{}.
\newblock \showarticletitle{Neural machine translation by jointly learning to
  align and translate}. In \bibinfo{booktitle}{\emph{ICLR'15}}.
\newblock


\bibitem[\protect\citeauthoryear{Chen, Ma, and Xiao}{Chen
  et~al\mbox{.}}{2018}]%
        {chen2018fastgcn}
\bibfield{author}{\bibinfo{person}{Jie Chen}, \bibinfo{person}{Tengfei Ma},
  {and} \bibinfo{person}{Cao Xiao}.} \bibinfo{year}{2018}\natexlab{}.
\newblock \showarticletitle{Fastgcn: fast learning with graph convolutional
  networks via importance sampling}. In \bibinfo{booktitle}{\emph{Proceedings
  of the 6th International Conference on Learning Representations}}.
\newblock


\bibitem[\protect\citeauthoryear{Dai, Dai, and Song}{Dai et~al\mbox{.}}{2016}]%
        {dai2016discriminative}
\bibfield{author}{\bibinfo{person}{Hanjun Dai}, \bibinfo{person}{Bo Dai}, {and}
  \bibinfo{person}{Le Song}.} \bibinfo{year}{2016}\natexlab{}.
\newblock \showarticletitle{Discriminative embeddings of latent variable models
  for structured data}. In \bibinfo{booktitle}{\emph{International conference
  on machine learning}}. \bibinfo{pages}{2702--2711}.
\newblock


\bibitem[\protect\citeauthoryear{Grover and Leskovec}{Grover and
  Leskovec}{2016}]%
        {grover2016node2vec}
\bibfield{author}{\bibinfo{person}{Aditya Grover} {and} \bibinfo{person}{Jure
  Leskovec}.} \bibinfo{year}{2016}\natexlab{}.
\newblock \showarticletitle{node2vec: Scalable feature learning for networks}.
  In \bibinfo{booktitle}{\emph{Proceedings of the 22nd ACM SIGKDD international
  conference on Knowledge discovery and data mining}}.
  \bibinfo{pages}{855--864}.
\newblock


\bibitem[\protect\citeauthoryear{Hamilton, Ying, and Leskovec}{Hamilton
  et~al\mbox{.}}{2017}]%
        {hamilton2017inductive}
\bibfield{author}{\bibinfo{person}{Will Hamilton}, \bibinfo{person}{Zhitao
  Ying}, {and} \bibinfo{person}{Jure Leskovec}.}
  \bibinfo{year}{2017}\natexlab{}.
\newblock \showarticletitle{Inductive representation learning on large graphs}.
  In \bibinfo{booktitle}{\emph{NeurIPS'17}}. \bibinfo{pages}{1024--1034}.
\newblock


\bibitem[\protect\citeauthoryear{Hu, Fey, Zitnik, Dong, Ren, Liu, Catasta, and
  Leskovec}{Hu et~al\mbox{.}}{2020b}]%
        {hu2020open}
\bibfield{author}{\bibinfo{person}{Weihua Hu}, \bibinfo{person}{Matthias Fey},
  \bibinfo{person}{Marinka Zitnik}, \bibinfo{person}{Yuxiao Dong},
  \bibinfo{person}{Hongyu Ren}, \bibinfo{person}{Bowen Liu},
  \bibinfo{person}{Michele Catasta}, {and} \bibinfo{person}{Jure Leskovec}.}
  \bibinfo{year}{2020}\natexlab{b}.
\newblock \showarticletitle{Open graph benchmark: Datasets for machine learning
  on graphs}. In \bibinfo{booktitle}{\emph{Advances in Neural Information
  Processing Systems}}.
\newblock


\bibitem[\protect\citeauthoryear{Hu, Dong, Wang, and Sun}{Hu
  et~al\mbox{.}}{2020a}]%
        {hu2020heterogeneous}
\bibfield{author}{\bibinfo{person}{Ziniu Hu}, \bibinfo{person}{Yuxiao Dong},
  \bibinfo{person}{Kuansan Wang}, {and} \bibinfo{person}{Yizhou Sun}.}
  \bibinfo{year}{2020}\natexlab{a}.
\newblock \showarticletitle{Heterogeneous graph transformer}. In
  \bibinfo{booktitle}{\emph{Proceedings of The Web Conference 2020}}.
  \bibinfo{pages}{2704--2710}.
\newblock


\bibitem[\protect\citeauthoryear{Huang, Zhang, Rong, and Huang}{Huang
  et~al\mbox{.}}{2018}]%
        {huang2018adaptive}
\bibfield{author}{\bibinfo{person}{Wenbing Huang}, \bibinfo{person}{Tong
  Zhang}, \bibinfo{person}{Yu Rong}, {and} \bibinfo{person}{Junzhou Huang}.}
  \bibinfo{year}{2018}\natexlab{}.
\newblock \showarticletitle{Adaptive sampling towards fast graph representation
  learning}. In \bibinfo{booktitle}{\emph{Advances in neural information
  processing systems}}, Vol.~\bibinfo{volume}{31}. \bibinfo{pages}{4558--4567}.
\newblock


\bibitem[\protect\citeauthoryear{Ji, Yin, Yang, Zhou, Zheng, Shi, and Fang}{Ji
  et~al\mbox{.}}{2020}]%
        {ji2020accelerating}
\bibfield{author}{\bibinfo{person}{Yugang Ji}, \bibinfo{person}{Mingyang Yin},
  \bibinfo{person}{Hongxia Yang}, \bibinfo{person}{Jingren Zhou},
  \bibinfo{person}{Vincent~W Zheng}, \bibinfo{person}{Chuan Shi}, {and}
  \bibinfo{person}{Yuan Fang}.} \bibinfo{year}{2020}\natexlab{}.
\newblock \showarticletitle{Accelerating Large-Scale Heterogeneous Interaction
  Graph Embedding Learning via Importance Sampling}.
\newblock \bibinfo{journal}{\emph{ACM Transactions on Knowledge Discovery from
  Data (TKDD)}} \bibinfo{volume}{15}, \bibinfo{number}{1}
  (\bibinfo{year}{2020}), \bibinfo{pages}{1--23}.
\newblock


\bibitem[\protect\citeauthoryear{Kipf and Welling}{Kipf and Welling}{2016}]%
        {kipf2016semi}
\bibfield{author}{\bibinfo{person}{Thomas~N Kipf} {and} \bibinfo{person}{Max
  Welling}.} \bibinfo{year}{2016}\natexlab{}.
\newblock \showarticletitle{Semi-supervised classification with graph
  convolutional networks}. In \bibinfo{booktitle}{\emph{International
  Conference on Learning Representations}}.
\newblock


\bibitem[\protect\citeauthoryear{Liao, He, Zhang, and Chua}{Liao
  et~al\mbox{.}}{2018}]%
        {liao2018attributed}
\bibfield{author}{\bibinfo{person}{Lizi Liao}, \bibinfo{person}{Xiangnan He},
  \bibinfo{person}{Hanwang Zhang}, {and} \bibinfo{person}{Tat-Seng Chua}.}
  \bibinfo{year}{2018}\natexlab{}.
\newblock \showarticletitle{Attributed social network embedding}.
\newblock \bibinfo{journal}{\emph{IEEE Transactions on Knowledge and Data
  Engineering}} \bibinfo{volume}{30}, \bibinfo{number}{12}
  (\bibinfo{year}{2018}), \bibinfo{pages}{2257--2270}.
\newblock


\bibitem[\protect\citeauthoryear{Liu, Chen, Li, Zhou, Li, Song, and Qi}{Liu
  et~al\mbox{.}}{2019}]%
        {liu2019geniepath}
\bibfield{author}{\bibinfo{person}{Ziqi Liu}, \bibinfo{person}{Chaochao Chen},
  \bibinfo{person}{Longfei Li}, \bibinfo{person}{Jun Zhou},
  \bibinfo{person}{Xiaolong Li}, \bibinfo{person}{Le Song}, {and}
  \bibinfo{person}{Yuan Qi}.} \bibinfo{year}{2019}\natexlab{}.
\newblock \showarticletitle{Geniepath: Graph neural networks with adaptive
  receptive paths}. In \bibinfo{booktitle}{\emph{Proceedings of the AAAI
  Conference on Artificial Intelligence}}, Vol.~\bibinfo{volume}{33}.
  \bibinfo{pages}{4424--4431}.
\newblock


\bibitem[\protect\citeauthoryear{Loshchilov and Hutter}{Loshchilov and
  Hutter}{2019}]%
        {loshchilov2017decoupled}
\bibfield{author}{\bibinfo{person}{Ilya Loshchilov} {and}
  \bibinfo{person}{Frank Hutter}.} \bibinfo{year}{2019}\natexlab{}.
\newblock \showarticletitle{Decoupled weight decay regularization}. In
  \bibinfo{booktitle}{\emph{ICLR'19}}.
\newblock


\bibitem[\protect\citeauthoryear{Luong, Pham, and Manning}{Luong
  et~al\mbox{.}}{2015}]%
        {luong2015effective}
\bibfield{author}{\bibinfo{person}{Minh-Thang Luong}, \bibinfo{person}{Hieu
  Pham}, {and} \bibinfo{person}{Christopher~D Manning}.}
  \bibinfo{year}{2015}\natexlab{}.
\newblock \showarticletitle{Effective approaches to attention-based neural
  machine translation}. In \bibinfo{booktitle}{\emph{Proceedings of the 2015
  conference on empirical methods in natural language processing}}.
  \bibinfo{pages}{1412--1421}.
\newblock


\bibitem[\protect\citeauthoryear{Mikolov, Sutskever, Chen, Corrado, and
  Dean}{Mikolov et~al\mbox{.}}{2013}]%
        {mikolov2013distributed}
\bibfield{author}{\bibinfo{person}{Tomas Mikolov}, \bibinfo{person}{Ilya
  Sutskever}, \bibinfo{person}{Kai Chen}, \bibinfo{person}{Greg~S Corrado},
  {and} \bibinfo{person}{Jeff Dean}.} \bibinfo{year}{2013}\natexlab{}.
\newblock \showarticletitle{Distributed representations of words and phrases
  and their compositionality}. In \bibinfo{booktitle}{\emph{Advances in neural
  information processing systems}}, Vol.~\bibinfo{volume}{26}.
  \bibinfo{pages}{3111--3119}.
\newblock


\bibitem[\protect\citeauthoryear{Paszke, Gross, Massa, Lerer, Bradbury, Chanan,
  Killeen, Lin, Gimelshein, Antiga, et~al\mbox{.}}{Paszke
  et~al\mbox{.}}{2019}]%
        {paszke2019pytorch}
\bibfield{author}{\bibinfo{person}{Adam Paszke}, \bibinfo{person}{Sam Gross},
  \bibinfo{person}{Francisco Massa}, \bibinfo{person}{Adam Lerer},
  \bibinfo{person}{James Bradbury}, \bibinfo{person}{Gregory Chanan},
  \bibinfo{person}{Trevor Killeen}, \bibinfo{person}{Zeming Lin},
  \bibinfo{person}{Natalia Gimelshein}, \bibinfo{person}{Luca Antiga},
  {et~al\mbox{.}}} \bibinfo{year}{2019}\natexlab{}.
\newblock \showarticletitle{Pytorch: An imperative style, high-performance deep
  learning library}. In \bibinfo{booktitle}{\emph{Advances in Neural
  Information Processing Systems}}.
\newblock


\bibitem[\protect\citeauthoryear{Pavlopoulos, Malakasiotis, and
  Androutsopoulos}{Pavlopoulos et~al\mbox{.}}{2017}]%
        {pavlopoulos2017deeper}
\bibfield{author}{\bibinfo{person}{John Pavlopoulos},
  \bibinfo{person}{Prodromos Malakasiotis}, {and} \bibinfo{person}{Ion
  Androutsopoulos}.} \bibinfo{year}{2017}\natexlab{}.
\newblock \showarticletitle{Deeper attention to abusive user content
  moderation}. In \bibinfo{booktitle}{\emph{Proceedings of the 2017 conference
  on empirical methods in natural language processing}}.
  \bibinfo{pages}{1125--1135}.
\newblock


\bibitem[\protect\citeauthoryear{Perozzi, Al-Rfou, and Skiena}{Perozzi
  et~al\mbox{.}}{2014}]%
        {perozzi2014deepwalk}
\bibfield{author}{\bibinfo{person}{Bryan Perozzi}, \bibinfo{person}{Rami
  Al-Rfou}, {and} \bibinfo{person}{Steven Skiena}.}
  \bibinfo{year}{2014}\natexlab{}.
\newblock \showarticletitle{Deepwalk: Online learning of social
  representations}. In \bibinfo{booktitle}{\emph{Proceedings of the 20th ACM
  SIGKDD international conference on Knowledge discovery and data mining}}.
  \bibinfo{pages}{701--710}.
\newblock


\bibitem[\protect\citeauthoryear{Vaswani, Shazeer, Parmar, Uszkoreit, Jones,
  Gomez, Kaiser, and Polosukhin}{Vaswani et~al\mbox{.}}{2017}]%
        {vaswani2017attention}
\bibfield{author}{\bibinfo{person}{Ashish Vaswani}, \bibinfo{person}{Noam
  Shazeer}, \bibinfo{person}{Niki Parmar}, \bibinfo{person}{Jakob Uszkoreit},
  \bibinfo{person}{Llion Jones}, \bibinfo{person}{Aidan~N Gomez},
  \bibinfo{person}{Lukasz Kaiser}, {and} \bibinfo{person}{Illia Polosukhin}.}
  \bibinfo{year}{2017}\natexlab{}.
\newblock \showarticletitle{Attention is all you need}. In
  \bibinfo{booktitle}{\emph{Advances in Neural Information Processing
  Systems}}.
\newblock


\bibitem[\protect\citeauthoryear{Veli{\v{c}}kovi{\'c}, Cucurull, Casanova,
  Romero, Lio, and Bengio}{Veli{\v{c}}kovi{\'c} et~al\mbox{.}}{2018}]%
        {velivckovic2017graph}
\bibfield{author}{\bibinfo{person}{Petar Veli{\v{c}}kovi{\'c}},
  \bibinfo{person}{Guillem Cucurull}, \bibinfo{person}{Arantxa Casanova},
  \bibinfo{person}{Adriana Romero}, \bibinfo{person}{Pietro Lio}, {and}
  \bibinfo{person}{Yoshua Bengio}.} \bibinfo{year}{2018}\natexlab{}.
\newblock \showarticletitle{Graph attention networks}. In
  \bibinfo{booktitle}{\emph{ICLR'18}}.
\newblock


\bibitem[\protect\citeauthoryear{Wang, Yu, Zheng, Gan, Gai, Ye, Li, Zhou,
  Huang, Ma, et~al\mbox{.}}{Wang et~al\mbox{.}}{2019b}]%
        {wang2019deep}
\bibfield{author}{\bibinfo{person}{Minjie Wang}, \bibinfo{person}{Lingfan Yu},
  \bibinfo{person}{Da Zheng}, \bibinfo{person}{Quan Gan}, \bibinfo{person}{Yu
  Gai}, \bibinfo{person}{Zihao Ye}, \bibinfo{person}{Mufei Li},
  \bibinfo{person}{Jinjing Zhou}, \bibinfo{person}{Qi Huang},
  \bibinfo{person}{Chao Ma}, {et~al\mbox{.}}} \bibinfo{year}{2019}\natexlab{b}.
\newblock \showarticletitle{Deep Graph Library: Towards Efficient and Scalable
  Deep Learning on Graphs}. In \bibinfo{booktitle}{\emph{ICLR'19}}.
\newblock


\bibitem[\protect\citeauthoryear{Wang, Ji, Shi, Wang, Ye, Cui, and Yu}{Wang
  et~al\mbox{.}}{2019a}]%
        {wang2019heterogeneous}
\bibfield{author}{\bibinfo{person}{Xiao Wang}, \bibinfo{person}{Houye Ji},
  \bibinfo{person}{Chuan Shi}, \bibinfo{person}{Bai Wang},
  \bibinfo{person}{Yanfang Ye}, \bibinfo{person}{Peng Cui}, {and}
  \bibinfo{person}{Philip~S Yu}.} \bibinfo{year}{2019}\natexlab{a}.
\newblock \showarticletitle{Heterogeneous graph attention network}. In
  \bibinfo{booktitle}{\emph{The World Wide Web Conference}}.
  \bibinfo{pages}{2022--2032}.
\newblock


\bibitem[\protect\citeauthoryear{Xu, Hu, Leskovec, and Jegelka}{Xu
  et~al\mbox{.}}{2018}]%
        {xu2018powerful}
\bibfield{author}{\bibinfo{person}{Keyulu Xu}, \bibinfo{person}{Weihua Hu},
  \bibinfo{person}{Jure Leskovec}, {and} \bibinfo{person}{Stefanie Jegelka}.}
  \bibinfo{year}{2018}\natexlab{}.
\newblock \showarticletitle{How Powerful are Graph Neural Networks?}. In
  \bibinfo{booktitle}{\emph{ICLR'18}}.
\newblock


\bibitem[\protect\citeauthoryear{Yang, Yang, Dyer, He, Smola, and Hovy}{Yang
  et~al\mbox{.}}{2016}]%
        {yang2016hierarchical}
\bibfield{author}{\bibinfo{person}{Zichao Yang}, \bibinfo{person}{Diyi Yang},
  \bibinfo{person}{Chris Dyer}, \bibinfo{person}{Xiaodong He},
  \bibinfo{person}{Alex Smola}, {and} \bibinfo{person}{Eduard Hovy}.}
  \bibinfo{year}{2016}\natexlab{}.
\newblock \showarticletitle{Hierarchical attention networks for document
  classification}. In \bibinfo{booktitle}{\emph{Proceedings of the 2016
  conference of the North American chapter of the association for computational
  linguistics: human language technologies}}. \bibinfo{pages}{1480--1489}.
\newblock


\bibitem[\protect\citeauthoryear{Ying, He, Chen, Eksombatchai, Hamilton, and
  Leskovec}{Ying et~al\mbox{.}}{2018}]%
        {ying2018graph}
\bibfield{author}{\bibinfo{person}{Rex Ying}, \bibinfo{person}{Ruining He},
  \bibinfo{person}{Kaifeng Chen}, \bibinfo{person}{Pong Eksombatchai},
  \bibinfo{person}{William~L Hamilton}, {and} \bibinfo{person}{Jure Leskovec}.}
  \bibinfo{year}{2018}\natexlab{}.
\newblock \showarticletitle{Graph convolutional neural networks for web-scale
  recommender systems}. In \bibinfo{booktitle}{\emph{Proceedings of the 24th
  ACM SIGKDD International Conference on Knowledge Discovery \& Data Mining}}.
  \bibinfo{pages}{974--983}.
\newblock


\bibitem[\protect\citeauthoryear{Yun, Jeong, Kim, Kang, and Kim}{Yun
  et~al\mbox{.}}{2019}]%
        {yun2019graph}
\bibfield{author}{\bibinfo{person}{Seongjun Yun}, \bibinfo{person}{Minbyul
  Jeong}, \bibinfo{person}{Raehyun Kim}, \bibinfo{person}{Jaewoo Kang}, {and}
  \bibinfo{person}{Hyunwoo~J Kim}.} \bibinfo{year}{2019}\natexlab{}.
\newblock \showarticletitle{Graph transformer networks}. In
  \bibinfo{booktitle}{\emph{Advances in Neural Information Processing
  Systems}}. \bibinfo{pages}{11983--11993}.
\newblock


\bibitem[\protect\citeauthoryear{Zhang, Song, Huang, Swami, and Chawla}{Zhang
  et~al\mbox{.}}{2019}]%
        {zhang2019heterogeneous}
\bibfield{author}{\bibinfo{person}{Chuxu Zhang}, \bibinfo{person}{Dongjin
  Song}, \bibinfo{person}{Chao Huang}, \bibinfo{person}{Ananthram Swami}, {and}
  \bibinfo{person}{Nitesh~V Chawla}.} \bibinfo{year}{2019}\natexlab{}.
\newblock \showarticletitle{Heterogeneous graph neural network}. In
  \bibinfo{booktitle}{\emph{Proceedings of the 25th ACM SIGKDD International
  Conference on Knowledge Discovery \& Data Mining}}.
  \bibinfo{pages}{793--803}.
\newblock


\bibitem[\protect\citeauthoryear{Zou, Hu, Wang, Jiang, Sun, and Gu}{Zou
  et~al\mbox{.}}{2019}]%
        {zou2019layer}
\bibfield{author}{\bibinfo{person}{Difan Zou}, \bibinfo{person}{Ziniu Hu},
  \bibinfo{person}{Yewen Wang}, \bibinfo{person}{Song Jiang},
  \bibinfo{person}{Yizhou Sun}, {and} \bibinfo{person}{Quanquan Gu}.}
  \bibinfo{year}{2019}\natexlab{}.
\newblock \showarticletitle{Layer-dependent importance sampling for training
  deep and large graph convolutional networks}. In
  \bibinfo{booktitle}{\emph{Advances in Neural Information Processing
  Systems}}. \bibinfo{pages}{11249--11259}.
\newblock


\end{thebibliography}



\end{document}